%% file: main.tex
\documentclass{article}

   \usepackage[preprint]{corl_2026} 

\usepackage{times}
\usepackage{helvet}
\usepackage{courier}
\usepackage[T1]{fontenc}
\usepackage[utf8]{inputenc}
\usepackage{microtype}
\usepackage{amsmath,amssymb,mathtools,bm}
\usepackage{booktabs,multirow,array}
\usepackage{graphicx,subcaption}
\usepackage{xcolor}
\usepackage{url}
\usepackage{hyperref}
\usepackage{cleveref}
\usepackage{algorithm,algorithmic}
\usepackage{enumitem}
\usepackage{listings}
\usepackage{xcolor}

\input{corl/sections/packages}

\title{\ours: Formally Verifiable Self-Evolving Skills for \\ Physical AI Agents}

\author{
  Yunhao Yang$^{*1}$ \, Neel P. Bhatt$^{*1}$ \, Kevin Wang$^{1}$ \, Samuel Tetteh$^{2}$ \,\\
  \textbf{Zhangyang Wang}$^{1}$ \, \textbf{Ufuk Topcu$^{1}$}\\
  $^1$The University of Texas at Austin \, $^2$Iowa State University \, $^*$Equal Contribution\\
  \texttt{\{yunhaoyang234,npbhatt,kevinwang.1839,atlaswang,utopcu\}@utexas.edu}\\
  \texttt{samtett@iastate.edu}
}

\begin{document}
\maketitle
\input{corl/sections/0_new_abstract}
\input{corl/sections/AW_new_intro}
\input{corl/sections/2_related_work}
\input{corl/sections/3_problem}
\input{corl/sections/4_method}
\input{corl/sections/5_experiment}
\input{corl/sections/6_conclusion}

\bibliography{references}

\input{corl/sections/7_appendix}

\end{document}

%% file: corl/sections/packages.tex
\usepackage{algorithm}
\usepackage{algorithmic}
\usepackage{enumitem}

\usepackage{wrapfig}
\usepackage{amsmath}

\lstdefinestyle{promptstyle}{
    basicstyle=\ttfamily\small,
    breaklines=true,
    frame=single,
    columns=fullflexible,
    keepspaces=true,
    showstringspaces=false,
    backgroundcolor=\color{gray!10},
    xleftmargin=2mm,
    xrightmargin=2mm
}

\definecolor{codegreen}{rgb}{0,0.6,0}
\definecolor{codegray}{rgb}{0.5,0.5,0.5}
\definecolor{codepurple}{rgb}{0.58,0,0.82}
\definecolor{backcolour}{rgb}{0.88,0.92,0.99}
\definecolor{figurepurple}{rgb}{0.631,0.333,0.906}
\definecolor{darkgreen}{RGB}{0,100,0}

\lstdefinestyle{code}{
    backgroundcolor=\color{backcolour},   
    commentstyle=\color{codegreen},
    keywordstyle=\color{magenta},
    numberstyle=\tiny\color{codegray},
    stringstyle=\color{codepurple},
    basicstyle=\ttfamily\footnotesize,
    breakatwhitespace=false,         
    breaklines=true,                 
    captionpos=b,                    
    keepspaces=true,                 
    numbers=left,                    
    numbersep=5pt,                  
    showspaces=false,                
    showstringspaces=false,
    showtabs=false,                  
    tabsize=2,
    moredelim=**[is][\color{blue}]{@}{@}
}

\usepackage{booktabs}
\usepackage{array}

\usepackage{xspace}
\newcommand{\ours}{\texttt{VASO}\xspace}

%% file: corl/sections/0_new_abstract.tex
\begin{abstract}


Reusable robot skills are becoming the basic units through which embodied agents turn open-ended instructions into long-horizon physical behavior. We argue that, while foundation models have collapsed the cost of creating these skills, \textit{the cost of trusting them has not}. Existing skill-evolution loops refine skills through execution feedback, unit tests, environment reward, or LLM self-critique, but these signals provide only trace-level evidence: they show that a skill worked on sampled executions, not that skill-induced plans satisfy temporal safety contracts under untested conditions. We introduce \ours, a framework for verification-guided self-evolution of LLM-generated robot skill contracts. In \ours, each skill is represented as a semantic contract with two coupled interfaces: a formal interface that aligns robot states, observations, and control commands with logical propositions for model checking, and a planner-facing interface that guides executable behavior generation. A model checker first filters logically inconsistent skill contracts, then verifies plans induced by the skill against global and local temporal specifications. When verification fails, \ours translates the counterexample trace into a textual gradient that updates the reusable skill contract while keeping foundation-model weights frozen. On Clearpath Jackal and PX4 quadcopter tasks, \ours reaches 97.2\% formal-specification compliance using fewer than 100 optimization samples, outperforming execution-feedback, prompt-optimization, and fine-tuning baselines. To our knowledge, \ours is the \textbf{first} framework that closes the loop between \textbf{formal verification} and \textbf{self-evolving} LLM-generated \textbf{skills} for \textbf{physical AI agents}: formal counterexamples become optimization feedback for reusable robot skill contracts, rather than merely verifying one-off plans, tuning planner prompts, or fine-tuning model weights.
\end{abstract}

%% file: corl/sections/AW_new_intro.tex
\section{Introduction}
\label{sec: intro}

Reusable skills are becoming the interface between foundation models and physical control. 
Instead of commanding a robot only through low-level actions, an embodied agent can invoke high-level modules such as navigation, object search, obstacle bypassing, or trajectory execution, and reuse them across long-horizon tasks under safety constraints~\cite{ahn2022can,liang2023code,Singh2022ProgPromptGS,Song2022LLMPlannerFG,CodeBotler}. 
Foundation models have made it increasingly easy to author such skills from natural language. 
A single prompt can now produce a new skill description, a code template, or a sequence of executable commands. 
However, lowering the cost of skill creation does not solve the harder problem of skill assurance. 
A generated skill can look reasonable in language while hiding a safety-critical failure in execution: a Jackal ground robot may continue moving after \texttt{person\_observed}, or a PX4 quadcopter may slightly exceed its altitude or velocity envelope. 
For physical AI agents, the central bottleneck is therefore no longer only how to create more skills, but how to make generated skills improve in a way that is inspectable, reusable, and verifiable.

Existing skill-library and self-evolving-agent methods close the skill-improvement loop primarily through soft feedback. 
Foundation-model-driven agents grow new skills from execution outcomes, unit tests, environment rewards, or LLM self-critique~\cite{wang2023voyager,wang2024trove,chen2024automanual,ma2024eureka,autoskill2026}. 
These feedback signals are powerful for expanding capability, because they tell an agent what worked on sampled tasks and allow successful behaviors to be reused. 
Yet they provide only trace-level evidence. 
They can show that a skill completed a task on observed executions, but they do not explain whether the same skill will induce plans that violate temporal specifications on untested cases. 
This distinction is crucial in robotics, where many successful rollouts do not compensate for a rare but severe violation of a safety rule. 
In our Jackal and PX4 benchmarks, even the strongest skill generated under execution-style feedback still leaves more than one plan in twenty violating its formal specification.

A natural idea is to bring \textbf{formal verification} into the skill-evolution loop. 
The difficulty is that a robot skill must serve several incompatible consumers at once. 
A model checker requires symbolic transition systems and temporal-logic propositions. 
A language-model planner requires a natural-language or code-like behavioral prior. 
A physical robot ultimately requires executable APIs, velocity commands, or trajectories. 
If verification is applied only after a plan has been generated, it becomes a final pass/fail filter rather than a mechanism for improving the reusable skill itself. 
If feedback is used only to tune a planner prompt or fine-tune a model, the learned improvement is not exposed as an interpretable skill artifact that can be inspected, reused, and verified independently of a single task instance. 
A \textbf{representation}, that lets formal counterexamples act directly on the skill, is missing.

We address this gap with a verifiable skill representation that couples a formal interface with a planner-facing interface. 
Each generated skill contains a temporal-logic rule, a proposition-aligned labeling function that maps robot states, observations, and control signals to logical propositions, and a text-based plan template that guides the planner. 
The labeling function is the bridge between the two worlds: it exposes skill-induced behavior to a model checker while keeping the skill usable by a language-model planner. 
This design makes it possible to treat verification not merely as an external safety check, but as an optimization signal for skill self-evolution.

\begin{figure*}[t]
    \centering
    \includegraphics[width=0.95\linewidth]{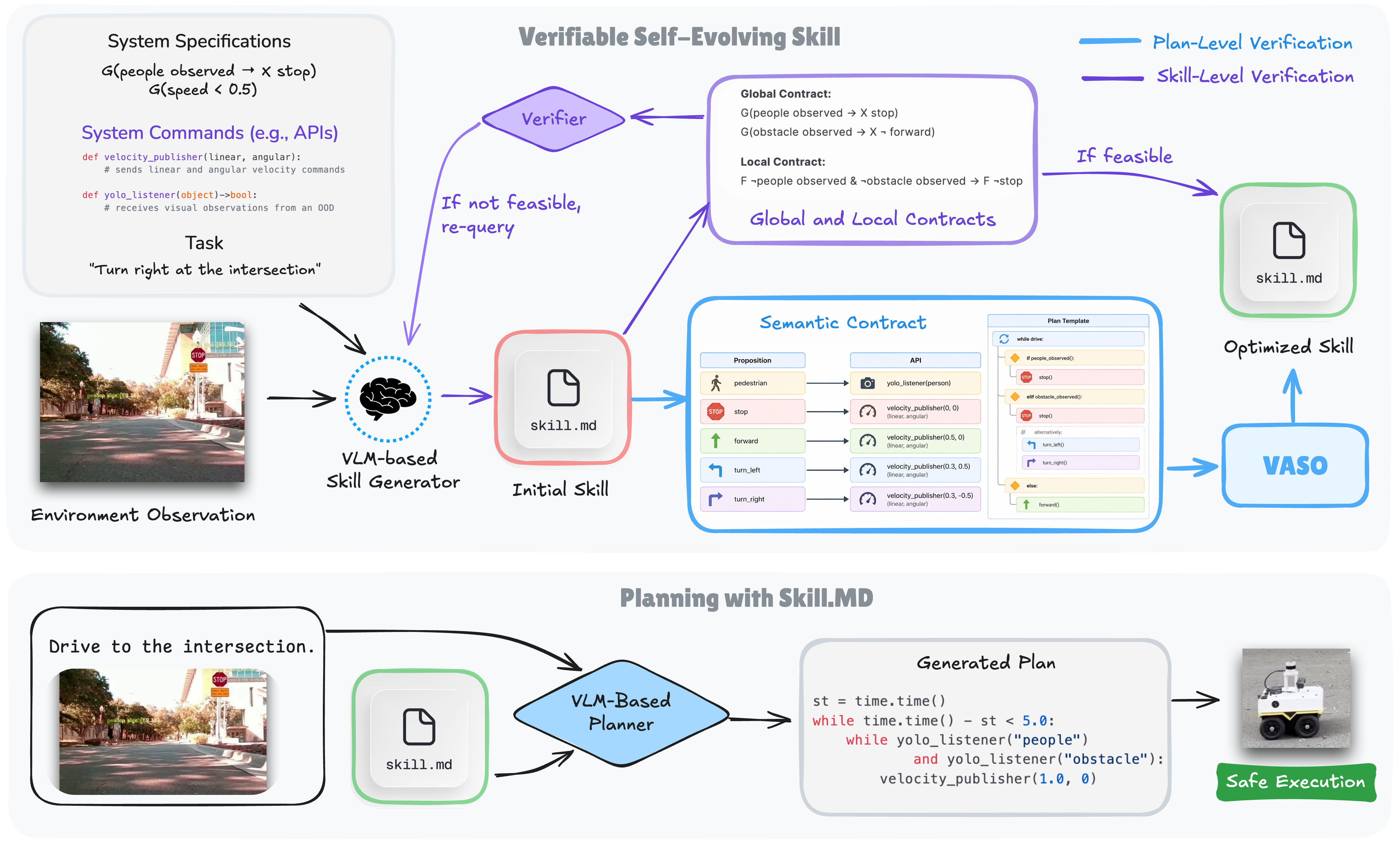}
    \caption{Overview of skill self-evolution loop. A task prompt yields a reusable skill, the skill is screened by skill-level feasibility checking, plans induced by the skill are verified against temporal specifications, and verifier counterexamples are translated into textual gradients that refine the skill.}
    \vspace{-12pt}
    \label{fig: pipeline}
\end{figure*}

We develop \ours (Verification-Guided Automated Skill Optimization), a closed-loop framework for verifiable self-evolving skills in physical AI agents. 
\ours operates at two levels. 
At the \textit{skill level}, it checks whether the generated local rule is logically consistent with the global specifications before any plan is sampled. 
This prevents the optimization loop from refining skills whose goals already contradict the safety rules. 
At the \textit{plan level}, \ours compiles an LLM-generated plan into a symbolic transition-system representation through the skill's proposition-aligned labeling function, then verifies the resulting behavior against global and local temporal specifications. 
When verification fails, the model checker returns a counterexample trace. 
\ours converts this counterexample into a textual gradient that updates the reusable skill, while keeping the foundation-model weights fixed. 
Thus, the object that evolves is not a one-off plan, a flat planner prompt, or hidden model parameters, but the skill representation itself.

To our knowledge, \ours is the \textbf{first} framework that closes the loop between formal verification and self-evolving LLM-generated skills for physical AI agents. Our framing also differs from standard verified planning pipelines: 
the goal is not only to \textit{reject} unsafe generated plans, but to use the reason for rejection to \textit{improve} the skill that produced them. 
As a result, the same refined skill can guide future plan generation with stronger specification compliance.

We evaluate \ours on two real robotic platforms, the Clearpath Jackal ground robot and a PX4 quadcopter, across $11$ temporal-logic specifications and $400$ generated plans. 
Optimized skills reach $97.2\%$ compliance with formal specifications using fewer than $100$ optimization samples and under $20$ minutes per skill, outperforming every zero-shot baseline as well as fine-tuning-based RLVF at a fraction of the training cost.

\noindent\textbf{Contributions:}
\textbf{(1) Verifiable self-evolving skill representation.} 
We introduce a reusable skill structure that connects temporal-logic specifications, proposition-aligned robot semantics, and planner-facing behavior templates, allowing LLM-generated skills to be both used for planning and exposed to formal verification.
\textbf{(2) Formal counterexamples as skill-optimization feedback.} 
We develop a two-level verification loop in which skill-level checking filters logically inconsistent rules, plan-level checking identifies behavioral violations, and counterexample traces are converted into textual gradients that refine the skill.
\textbf{(3) Physical-agent validation.} 
On real robot platforms, \ours reaches $97.2\%$ specification compliance under $100$ optimization samples, outperforming fine-tuning baselines while freezing the model parameters.

%% file: corl/sections/2_related_work.tex
\section{Related Work}
\label{sec: related-work}

\textbf{LLM Planning and Plan Verification.}
LLM planning methods generate executable action sequences~\cite{CodeBotler,Song2022LLMPlannerFG}, code policies~\cite{liang2023code,Singh2022ProgPromptGS}, solver aided formulations~\cite{Liu2023LLMPEL,wang2024llm}, and reflection based reasoning traces~\cite{wei2022chain,yao2022react,shinn2024reflexion}. Related systems also select from fixed skill sets via affordance grounding~\cite{ahn2022can}. In parallel, verification work compiles LLM outputs into automata or transition systems for model checking~\cite{liu2023grounding,grigorev2025verifyllm,hao2025large}, with related efforts adding uncertainty calibrated guarantees and multimodal grounding~\cite{bhattknow,yang2024aamas}. These methods remain at the plan level. Each task generates or verifies a fresh candidate, and counterexamples mainly serve as rejection certificates. \ours turns the same counterexample into a reusable update signal for the skill that produced the failed plan.

\textbf{Prompt and Model Optimization.}
Prompt optimization methods refine model inputs through textual gradients and programmatic prompt tuning~\cite{yuksekgonul2024textgrad,yin2025llm-autodiff,khattab2024dspy}, while LAD-VF routes verifier counterexamples back as prompt gradients~\cite{yang2025ladvf}. Model optimization instead changes weights through formal feedback~\cite{yang-mlsys}, preferences~\cite{RLHF,dpo}, or simulator data~\cite{hu2024robo}. These methods establish that feedback can improve generation, but they absorb the correction into a prompt or model parameters. \ours moves the correction to the reusable skill contract, making one verifier counterexample influence future plans induced by the same skill.

\textbf{Skill Evolution.}
Foundation model driven agents grow reusable skill libraries through open ended exploration~\cite{wang2023voyager,wang2024trove}, interactive manuals and self critique~\cite{chen2024automanual,autoskill2026}, or reward feedback~\cite{ma2024eureka, yang2026skillopt}. These works optimize skills, but their supervision is still at a trace level. A successful rollout shows that a skill worked on sampled executions, while temporal safety violations can remain outside the sampled set. \ours adds the missing formal supervision. The evolving object is the reusable skill contract itself, updated by the verifier counterexample that exposes its failure mode.

%% file: corl/sections/3_problem.tex
\section{Problem Formulation}
\label{sec: problem}

We consider an embodied agent operating over a sequence of low-level behaviors, such as API calls or waypoint trajectories. We define a set $\mathcal{AP} = \{a_1, \dots, a_n\}$ of atomic propositions.
A system execution is represented as a trace $\sigma = \sigma_0, \sigma_1, \sigma_2, \dots$, where each \(\sigma_t\) denotes an execution state at time \(t\). For ease of reference, we provide a \emph{table of notation} in Tab. \ref{tab:notations} in the appendix.

\paragraph{Skill Definition.}
We formally define a skill \(sk\) as a tuple $(\mathcal{G}, \psi_{sk}, \mathcal{C}_{sk})$, where \(\mathcal{G} = \{\varphi_1, \dots, \varphi_m\}\) is a set of global specifications given as LTL formulas over \(\mathcal{AP}\), and \(\psi_{sk}\) is a local rule given as an LTL formula over \(\mathcal{AP}\).

The semantic contract $\mathcal{C}_{sk} = (L_{sk}, \mathcal{C}_{sk}^{p})$, which is the core of a skill, consists of a proposition-aligned labeling function \(L_{sk}\) and \(\mathcal{C}_{sk}^{p}\) is a text-based plan template.
The labeling function $L_{sk} : \sigma_t \rightarrow 2^{\mathcal{AP}}$ maps system executions $\sigma_t$ to the set of atomic propositions that hold at time \(t\).
For an execution trace $\sigma = \sigma_0, \sigma_1, \dots$, the labeling function induces a proposition trace $L_{sk}(\sigma_0), L_{sk}(\sigma_1), \dots$.
The plan template \(\mathcal{C}_{sk}^{p}\) defines a semantic behavioral prior that guides plan generation. 

The global specifications define constraints that must always be satisfied:
\[
L_{sk}(\sigma) \models \mathcal{G} \quad \text{if and only if} \quad\forall i,\; L_{sk}(\sigma) \models \varphi_i.
\]
The local rule \(\psi_{sk}\) specifies the intended behavior or goal of the particular skill.

\paragraph{Planning.}
Given a task prompt \(\tau\) and a set \(\mathcal{S}\) of system commands (e.g., APIs), a foundation model $\mathcal{F}_{\text{skill}}$ generates a skill $sk = \mathcal{F}_{\text{skill}}(\tau,\mathcal{S},\mathcal{G})$.
The semantic contract \(\mathcal{C}_{sk}\) is then sent to a planner $\mathcal{F}_{\text{plan}}$ and used to generate a plan $\pi = \mathcal{F}_{\text{plan}}(\mathcal{C}_{sk}, \tau)$. The generated plan induces a set of possible executions represented by execution traces \(\sigma\), and the labeling function \(L_{sk}\) induces the corresponding proposition trace used for verification.

\paragraph{Plan Verification.}
We develop a an algorithm that compiles a plan into a transition system $\mathcal{A}_{\pi} = (S, S_0, T, L)$,
where each state \(s_t \in S\) corresponds to an execution state \(\sigma_t\), and each transition $(s_t, s_{t+1}) \in T$ represents an executable behavior transition induced by the plan \(\pi\), such as an API call or waypoint update. The labeling function $L(s_t) = L_{sk}(\sigma_t)$ maps each state to the atomic propositions satisfied by the corresponding execution state. We present the detailed algorithm in Alg. \ref{alg:plan-to-automaton} and examples in Tab. \ref{tab:nusmv_python} in the Appendix.

Then, we can verify the plan against the specifications via a formal verifier:
\[
\mathcal{A}_{\pi} \otimes \mathcal{M}
\models
\mathcal{G} \wedge \psi_{sk},
\]
where \(\mathcal{M}\) is a predefined transition system representing the environment and system dynamics over all possible executions.

\paragraph{Problem.}
Given atomic propositions \(\mathcal{AP}\), system commands \(\mathcal{S}\), global specifications \(\mathcal{G}\), and a task prompt \(\tau\), construct a \emph{self-evolving skill} $sk = (\mathcal{G}, \psi_{sk}, \mathcal{C}_{sk})$ such that it is logically \emph{feasible} (local rule does not contradict to the global specifications):
$
\exists \sigma \quad L_{sk}(\sigma) \models \mathcal{G} \wedge \psi_{sk},
$
and the generated plan satisfies the specifications under all possible executions: $\mathcal{A}_{\pi} \otimes \mathcal{M} \models \mathcal{G} \wedge \psi_{sk}.$

%% file: corl/sections/4_method.tex
\section{Verifiable Self-Evolving Skills}

Given a task prompt \(\tau\), a set of executable system commands \(\mathcal{S}\) (e.g., APIs), and a set of specifications \(\mathcal{G}\), we construct \emph{verifiable self-evolving skills} for plan generation. 
Specifically, we use a foundation model \(\mathcal{F}_{\text{skill}}\) to construct a skill
$
sk =(\mathcal{G},\psi_{sk},\mathcal{C}_{sk})
=
\mathcal{F}_{\text{skill}}(\tau,\mathcal{S},\mathcal{G}),
$
where \(\psi_{sk}\) is a local rule and 
\(\mathcal{C}_{sk}=(L_{sk},\mathcal{C}_{sk}^{p})\) is a semantic contract.

During optimization, we preserve the global specifications \(\mathcal{G}\) and iteratively \textbf{refine \(\psi_{sk}\) and \(\mathcal{C}_{sk}\) through verification feedback}. This procedure fixes foundation model weights and treats generated plans as \textbf{evidence for skill refinement}. 
Fig.~\ref{fig: pipeline} illustrates the framework with a running example.

\subsection{Skill-Level Verification and Local Rule Refinement}
\label{sec:verification}

The first verification stage checks whether the generated skill specification is logically feasible, i.e., whether the local rule is consistent with the global specifications.

Given the skill $sk = (\mathcal{G}, \psi_{sk}, \mathcal{C}_{sk})$, we define a combined specification
$\Phi_{sk} = \left( \bigwedge_{\varphi \in \mathcal{G}} \varphi \right) \wedge \psi_{sk}.$
We verify whether there exists an execution trace satisfying the combined specification: $\exists \sigma \quad \sigma \models \Phi_{sk}$.

To perform this check, we construct a universal transition system $\mathcal{M} = (S, S_0, T, L)$,
which represents all possible proposition transitions over \(\mathcal{AP}\). Specifically, \(S = 2^{\mathcal{AP}}, S_0 = S, T = S \times S \text{, and } L(s) = s, \ \forall s \in 2^{\mathcal{AP}} \).
Under this construction, feasibility checking reduces to satisfiability checking over all possible executions.
Instead of directly searching for a satisfying trace, we equivalently verify $\mathcal{M} \models \neg \Phi_{sk}$ via a model checker.

If the property holds, then all possible traces violate \(\Phi_{sk}\), implying that no satisfying execution exists and the skill is \emph{infeasible}. Then, we will re-query $\mathcal{F}_{skill}$ to refine the local rule, which we present an example in Listing \ref{lst:regen} in Appendix \ref{app: case}. 
Otherwise, the model checker returns a counterexample trace $\sigma \models \Phi_{sk},$ which serves as a witness that the skill specification is logically \emph{feasible}.

This skill-level verification checks logical feasibility of the generated specification rather than behavioral correctness. By filtering out logically inconsistent skills early, the framework avoids optimization over contradictory specifications and reduces downstream verification failures.

\begin{figure*}[t]
    \centering
    \includegraphics[width=0.95\linewidth]{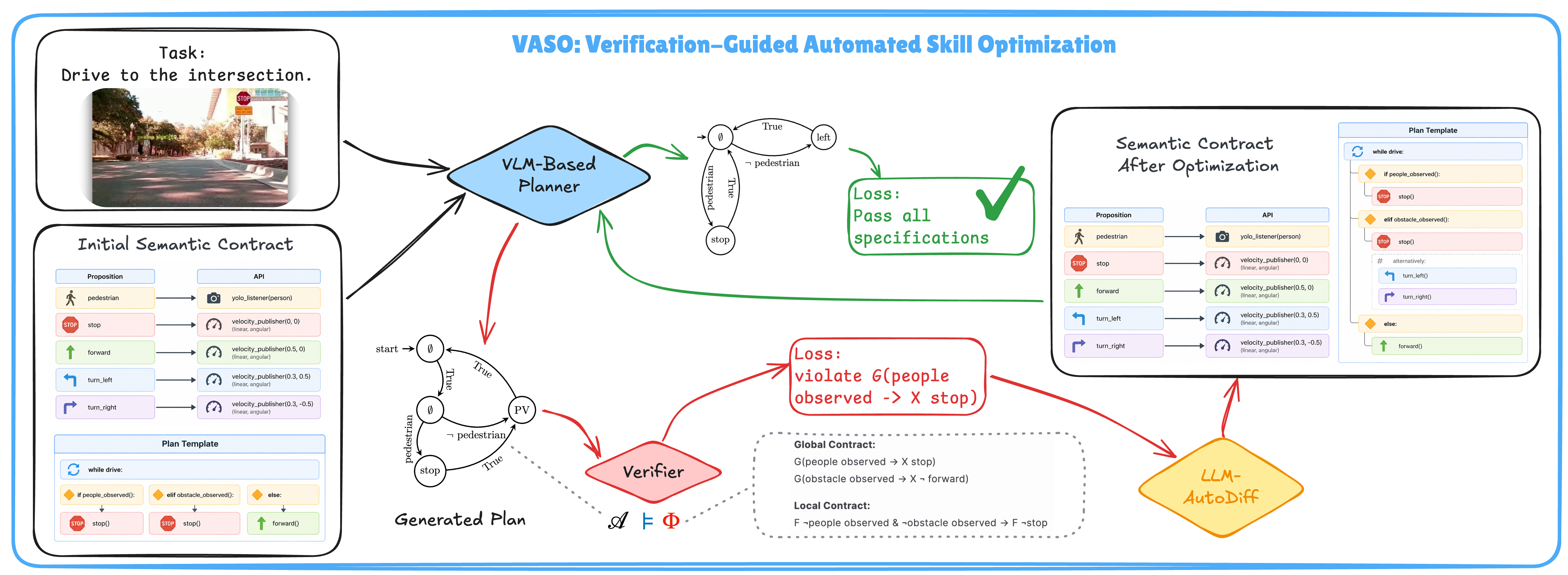} 
    \caption{An overview of \ours. It verifies plans generated from the semantic contract, and uses the feedback to iteratively refine the contract in a closed loop.}
    \label{fig: vaso}
\end{figure*}

\subsection{Verification-Guided Automated Skill Optimization (\ours)}

\paragraph{Plan-Level Behavior Verification.}

After obtaining a logically feasible skill, we generate a plan $\pi = \mathcal{F}_{\text{plan}}(\mathcal{C}_{sk}, \tau)$, where the semantic contract \(\mathcal{C}_{sk}\) guides plan synthesis. Then, we compile the plan into an automaton $\mathcal{A}_{\pi} = (S, S_0, T, L)$ as described in Sec. \ref{sec: problem} (Plan Verification).

Instead of directly reasoning over symbolic propositions, we derive symbolic states through the proposition-aligned labeling function $L_{sk}$. Specifically, each execution state is mapped to the set of propositions satisfied at that timestep: $L(s_t) = L_{sk}(\sigma_t).$
Applying \(L_{sk}\) over the execution sequence induces the proposition trace used for verification.
We then formally verify: $\mathcal{A}_{\pi} \models \mathcal{G} \wedge \psi_{sk}.$ 

Implementation-wise, the planner generates executable plans in the form of API calls or waypoint trajectories. The generated plans are then compiled into automata expressed in \textsc{NuSMV} \cite{Cimatti2002NuSMV} for formal verification. We provide examples in Tab. \ref{tab:nusmv_python} in the Appendix. If verification fails, the model checker returns a counterexample trace identifying a specification violation. 

\paragraph{Guarantee and Assumption.}
Due to the auto-generated proposition-aligned labeling function, the plan-level verification provides a \textbf{guarantee conditioned on the correctness of proposition alignments}. The goal of this verification step is to \textbf{automatically produce feedback} later used for semantic contract refinement.

\paragraph{Semantic Contract Refinement.}
We formulate semantic contract refinement as an iterative prompt optimization problem. Given a task prompt \(\tau\) and a skill $sk = (\mathcal{G}, \psi_{sk}, \mathcal{C}_{sk})$, the planner generates an executable plan: $\pi = \mathcal{F}_{\text{plan}}(\mathcal{C}_{sk}, \tau)$.

\begin{wrapfigure}{r}{0.5\textwidth}
    \centering
    \vspace{-20pt}
    \includegraphics[page=1,width=\linewidth]{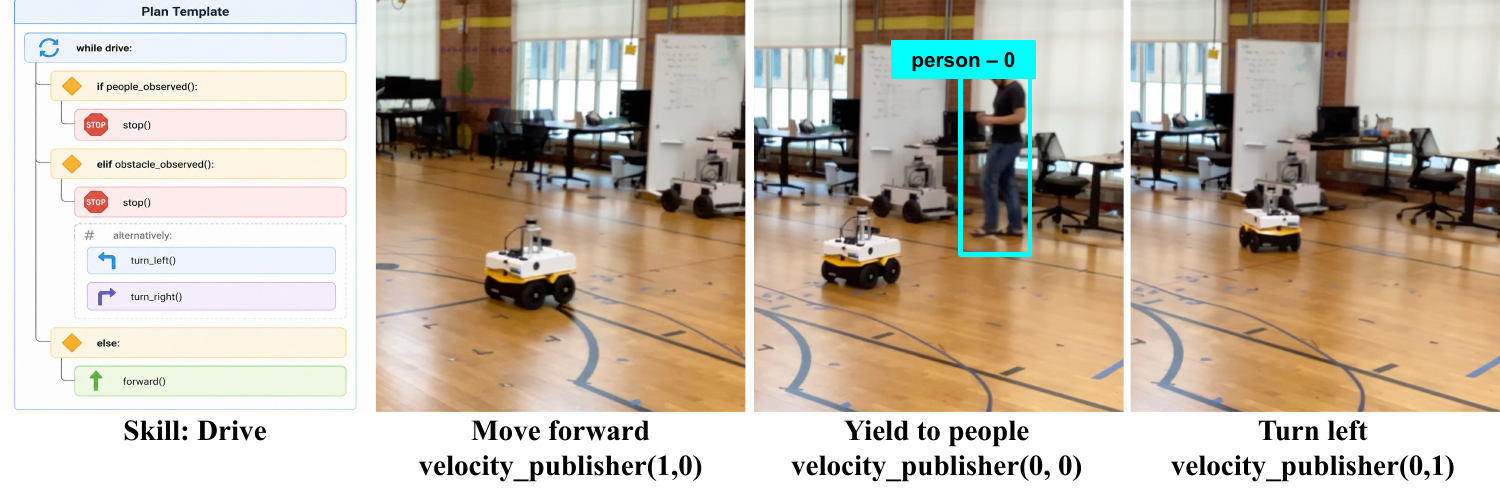}
    \includegraphics[page=2,width=\linewidth]{corl/figures/jackal-exe.pdf}
    \includegraphics[page=3,width=\linewidth]{corl/figures/jackal-exe.pdf}
    \caption{Real-world executions on the Jackal robot with multiple verifiable self-evolving skills.}
    \label{fig:jackal-exe}
\end{wrapfigure}

Our goal is to optimize the semantic contract such that the generated plan satisfies the specifications:
\[
\mathcal{C}_{sk}^{*} = \arg\min_{\mathcal{C}_{sk}} \mathcal{L} \big( \mathcal{F}_{\text{plan}}(\mathcal{C}_{sk}, \tau) \big),
\]
where \(\mathcal{L}\) is a loss that measures the percentage of violated specifications over all specifications.

Given a generated plan \(\pi\), the verifier returns:
\[ 
y \in \{ \texttt{pass}, \texttt{fail}(\varphi) \}, \quad \varphi \in \mathcal{G} \cup \psi_{sk}.
\]

We convert the verification outcome into a \textbf{textual gradient} \(g\), e.g., \textit{\color{purple}``violate specification: G(person observed -> X stop)''} or \textit{\color{darkgreen}``pass all specifications.''}
This textual gradient serves as a discrete approximation of $\frac{\partial \mathcal{L}}{\partial \mathcal{C}_s}$, guiding how the semantic contract should be refined to reduce specification violations.

We iteratively refine the semantic contract using the generated feedback: 
\[
\mathcal{C}_{sk}^{(k+1)} = \mathcal{F}_{\text{skill}} \big( \mathcal{C}_{sk}^{(k)}, g, \tau \big),
\]
where $k$ is the number of iterations. The textual gradient serves as natural-language feedback that describes the specification violations induced by the semantic contract. The updated semantic contract is then used to regenerate plans in the next optimization iteration. Hence, the optimization alternates between:
\[
\text{(Forward)} :\quad \mathcal{C}_{sk}^{(k)} \rightarrow \pi^{(k)} \rightarrow \mathcal{A}_{\pi^{(k)}},
\quad \text{and} \quad
\text{(Backward)} :\quad \mathcal{A}_{\pi^{(k)}} \rightarrow g^{(k)} \rightarrow \mathcal{C}_{sk}^{(k+1)}.
\]
This process can be interpreted as a form of textual gradient descent, where the verifier provides a structured optimization signal and the model $\mathcal{F}_{skill}$ performs the semantic contract update.
Implementation-wise, we adapt \texttt{LLM-AutoDiff} \cite{yin2025llm-autodiff} with default parameters. We show a illustration with a running example in Fig. \ref{fig: vaso}.

Importantly, the optimization operates entirely at the prompt level and does not require parameter updates, gradient backpropagation, or human supervision. It is fully automated through verification feedback, enabling efficient semantic contract optimization without human in the loop.

%% file: corl/sections/5_experiment.tex
\section{Empirical Evaluation}

We evaluate the proposed framework on two robotic platforms: the \texttt{Clearpath Jackal} ground robot and a \texttt{PX4} quadrotor drone. We use \texttt{GPT-5-nano} as $\mathcal{F}_{\text{skill}}$ and \texttt{GPT-4o-mini} as $\mathcal{F}_{\text{plan}}$. The Jackal is controlled through APIs for velocity control and object detection (see Fig. \ref{fig: pipeline}), while the drone executes velocity trajectories represented in the NED frame (see Fig. \ref{fig: execution-example}). Detailed platform-specific case studies are provided in Appendix~\ref{app: case}.

The evaluation addresses four questions: \textbf{(i)} Does verification feedback improve skills? \textbf{(ii)} Does optimizing a skill outperform flat prompt optimization on natural language prompts? \textbf{(iii)} Can optimized skills generalize to unseen task prompts? \textbf{(iv)} Does automatic proposition alignment provide a practical alternative to handcrafted mappings?

\begin{wrapfigure}{r}{0.6\textwidth}
    \centering
    \vspace{-5pt}
    \includegraphics[width=\linewidth]{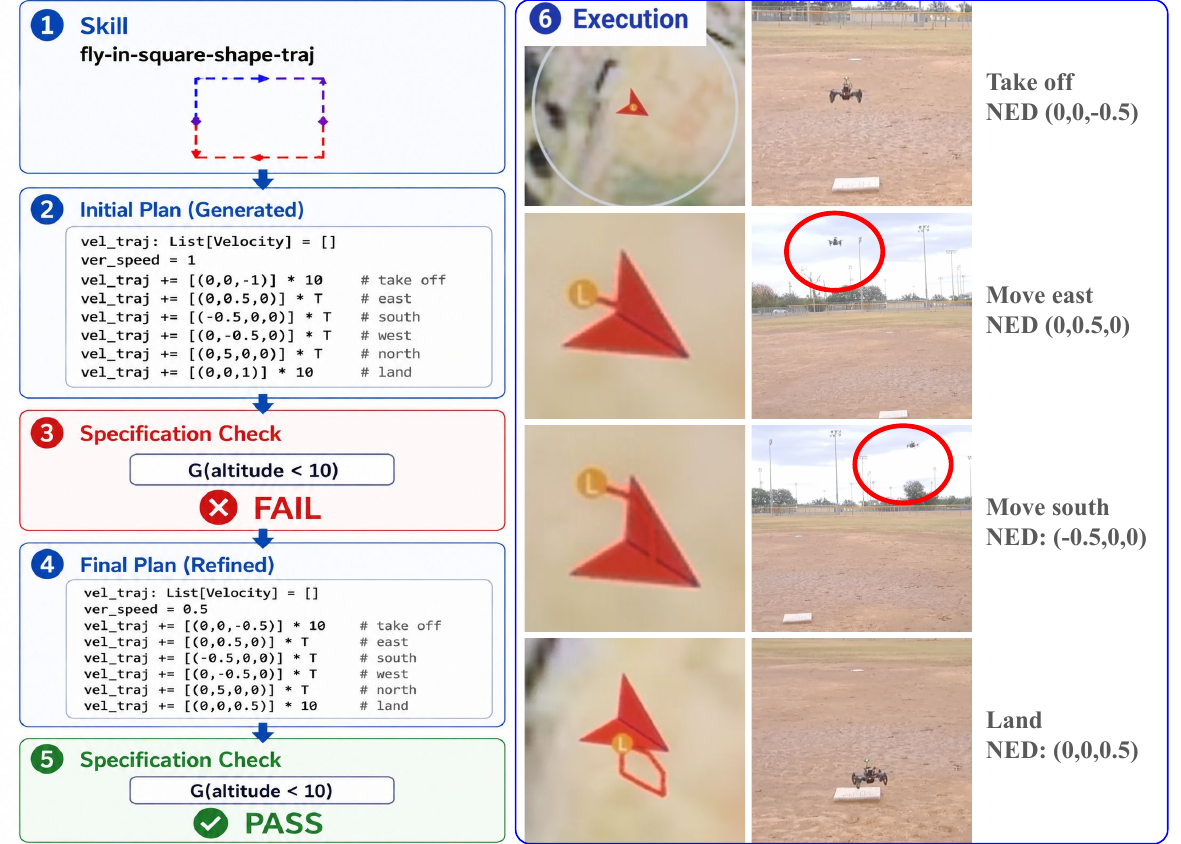}
    \caption{A running example of a self-evolving skill and plan executions on the PX4 drone, where the plans are velocity trajectories in NED frames. 
    }
    \vspace{-20pt}
    \label{fig: execution-example}
\end{wrapfigure}

\paragraph{Skill Evolving via Verification Feedback.}
We evaluate the framework on 11 temporal logic specifications (Listing \ref{lst:spec} in the Appendix) and 20 generated skills. 
Starting from raw outputs of $\mathcal{F}_{\text{skill}}$, \textbf{89\%} of generated skills satisfy all contracts. A failure example is shown in Listing~\ref{lst:regen}. With verification feedback, feasibility improves to \textbf{97\%} after one iteration, demonstrating rapid convergence with minimal re-queries.

Then, we measure skill quality by defining a safety score 
\[
\text{Safety}(\pi) = \frac{1}{N} \sum_{i=1}^N \mathbf{1}[\pi \models \varphi_i],
\]
which is the fraction of specifications satisfied by a generated plan.

\begin{figure}[t]
    \centering
    \includegraphics[height=0.24\linewidth]{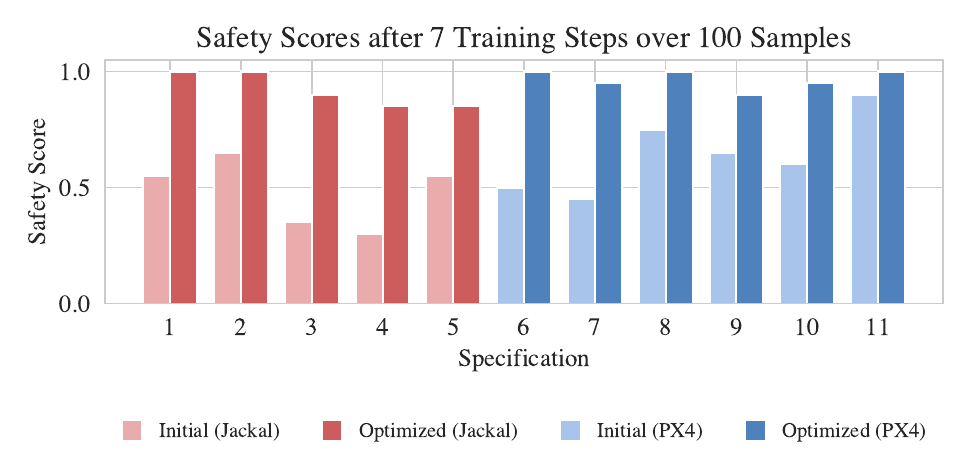}
    \includegraphics[height=0.24\linewidth]{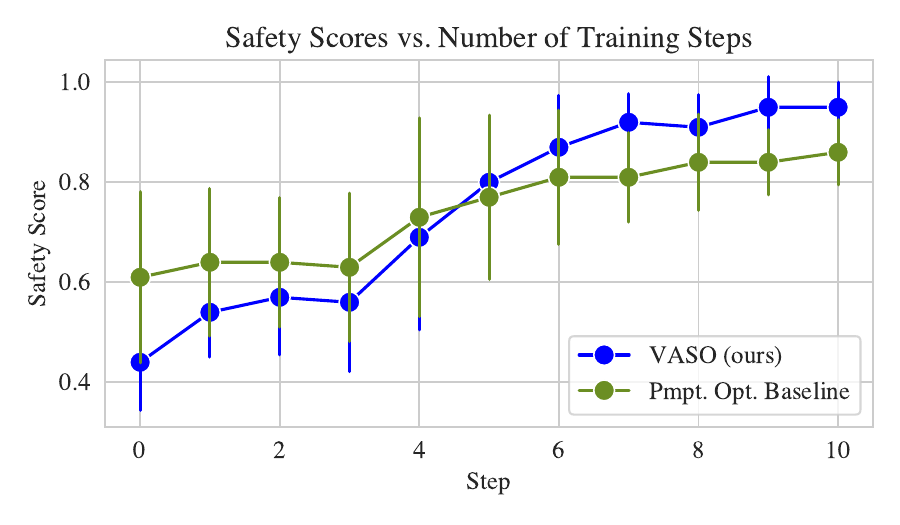}
    \caption{The left plot shows the safety score across 11 specifications. The right plot compares \ours with a baseline prompt optimization method that does not use structured skill representations.}
    \vspace{-10pt}
    \label{fig: testing}
\end{figure}

Fig. \ref{fig: testing} (left) compares 100 generated plans derived from the initial skills against 100 from optimized skills (after 7 steps) across all 11 specifications and both platforms. Optimized skills achieve about \textbf{95\%} safety scores, substantially outperforming the initial skills, while maintaining high task feasibility (97\%), indicating that verification feedback improves skill quality rather than merely encouraging conservative safe behavior.

\paragraph{Skill Representation vs. Flat Prompt.}
We compare \ours against flat prompt optimization, \texttt{LAD-VF}~\cite{yang2025ladvf}, that directly optimize natural-language instructions without an explicit skill representation. \texttt{LAD-VF} applies textual gradient from verification feedback to updates the prompts. Fig.~\ref{fig: testing} (right) shows that \ours exceeds \textbf{90\%} safety within \textbf{7 steps}, while \texttt{LAD-VF}, which lacks a structured skill abstraction, plateaus near \textbf{85\%} even after \textbf{10 steps}. It suggests that optimizing a skill provides a more effective optimization target than directly refining natural language prompts.

\begin{figure}[t]
    \centering
    \includegraphics[height=0.27\linewidth]{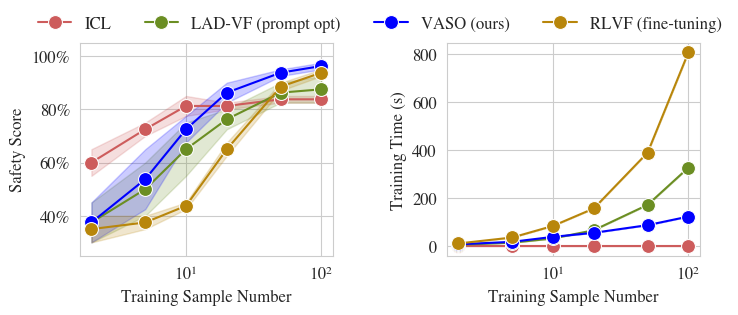}
    \includegraphics[height=0.27\linewidth]{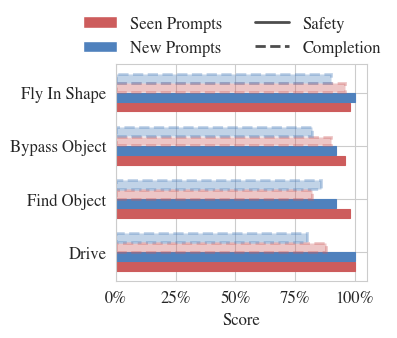}
    \caption{The left two figures compares the safety scores and training time of \ours against baselines. The right figure compares the safety and task completion scores on seen and unseen task prompts.}
    \vspace{-10pt}
    \label{fig: acc-size}
\end{figure}

\paragraph{Held-Out Skill Reuse.}
We select four representatives to evaluate the reuse capability of optimized skills to unseen task prompts. For each skill, we evaluate 5 seen prompts used during optimization and 5 new prompts from the same task family (details in App.~\ref{app:heldout-prompts}). As shown in Fig.~\ref{fig: acc-size} (right), \ours achieves \textbf{92--100\%} safety and \textbf{80--96\%} task completion on new prompts, suggesting that optimized skills can be reused without additional refinement.

\paragraph{Practicality of Automatic Proposition Alignment.}
A key practical challenge is constructing proposition-aligned labeling functions that map low-level robot controls to symbolic propositions. Instead of requiring manual engineering, we query $\mathcal{F}_{\text{skill}}$ to automatically generate proposition alignment functions from APIs and specifications. Since these mappings may contain abstraction errors, we compare them against handcrafted proposition alignment functions.

As shown in Tab.~\ref{tab:planning-baselines}, automatically generated proposition alignment achieves \textbf{on-par performance} with manual alignment while eliminating the need for handcrafted labeling functions, suggesting that automatic proposition alignment makes verification-guided skill optimization practical without requiring domain experts to manually define low-level symbolic abstractions.

\paragraph{Planning Baseline Comparison.}
We compare \ours against zero-shot planners and learning-based methods on safety score (SS) and task completion rate (TC, evaluated in simulation). As shown in Tab.~\ref{tab:planning-baselines}, \ours achieves the highest SS (\textbf{96.6\%}) while maintaining competitive TC. Compared to learning-based baselines, \ours attains a stronger safety-efficiency trade-off without parameter fine-tuning. Baseline details are provided in App.~\ref{app:experiment}.

We further compare the efficiency-performance trade-off against a few selected representatives: \texttt{LAD-VF} \cite{yang2025ladvf} (flat prompt optimization), \texttt{RLVF}~\cite{yang-mlsys} (fine-tuning foundation models via verification feedback), and \texttt{ICL}~\cite{dong2022survey} (handcrafted in-context examples satisfying all specifications). As shown in Fig.~\ref{fig: acc-size}, with only \textbf{100} training samples, \ours achieves the \textbf{highest safety score} while requiring only \textbf{one-fourth the training time} of fine-tuning approaches.

\begin{table}[t]
\centering
\caption{Planning baseline comparison on safety score (SS) and task completion (TC) rate.}
\label{tab:planning-baselines}

\setlength{\tabcolsep}{4pt}
\renewcommand{\arraystretch}{1.0}

\begin{tabular}{lcc|lcc|lcc}
\toprule

Zero-shot & SS & TC & Learning-based & SS & TC  & \textbf{Ours} & SS & TC\\
\midrule



LLM+P \cite{Liu2023LLMPEL} & 79.5 & 88.3 &
DSPy \cite{khattab2024dspy} & 86.8 & 87.0 & handcraft prop align & 97.2 & 85.3\\

ReAct \cite{react-agent} & 73.3 & 86.5 &
RoboInstruct \cite{hu2024robo} & 82.5 & 67.0 & automatic prop align & \textbf{96.8} & \textbf{86.5}\\

LLM-Planner \cite{Song2022LLMPlannerFG} & 90.3 & 89.3 &
RLVF \cite{yang-mlsys} & 94.3 & 82.5 & & &\\
\bottomrule
\end{tabular}

\vspace{-2mm}
\end{table}

%% file: corl/sections/6_conclusion.tex
\vspace{-8pt}
\section{Conclusion}
\vspace{-8pt}
We presented an automatic skill optimization framework for foundation model planning. By introducing a structured skill representation, our approach treats skills as both interpretable planning primitives and verifiable units of behavior.
Building on this representation, we developed a closed-loop pipeline that verifies the plans derived from the skill and automatically refines the skill via the verification feedback.
Empirical results show that \ours significantly outperforms existing LLM-based planning baselines and LLM fine-tuning approaches in both compliance and convergence efficiency. 

\vspace{-8pt}
\section{Limitations and Future Directions}
\vspace{-8pt}
The proposed framework has the following limitations:
\textbf{(i)} The proposition-aligned labeling function $L_{sk}$ is AI-generated and unverified, hence, errors in the proposition mapping can invalidate formal guarantees. 
\textbf{(ii)} Skills are verified under sequential, non-concurrent execution. The framework do not extend to overlapping or interleaved skill executions.
As future directions, we aim to improve the reliability of proposition-aligned labeling by incorporating uncertainty signals such as model confidence calibration. Another direction is to extend the framework to multi-agent settings where multiple skills interact simultaneously under shared environmental constraints.

%% file: corl/sections/7_appendix.tex
\newpage
\appendix

\section{Additional Implementation Details}

For skill generation, we synthesize a prompt to guide the skill generation model $\mathcal{F}_{skill}$, which we present in Listing \ref{lst:skill-prompt}.

\begin{lstlisting}[language={}, caption={A prompt template for skill generation.}, label={lst:skill-prompt}]
Given:

1. A set of low-level APIs or control commands
2. A set of global logical specifications

derive a robotic skill specification with the following sections:

1. Skill Name
2. Global Contract
3. Local Contract
4. Semantic Contract

Requirements:

- The Global Contract must preserve the provided global logical specifications.
- The Local Contract must define skill-specific logical behavior using temporal logic.
- The Semantic Contract must bridge atomic propositions and executable Python code.

For the Semantic Contract:

- Generate proposition-aligned Python APIs whose names exactly match the atomic propositions.
- Observation propositions should map to boolean-returning functions.
- Action propositions should map to executable control functions.
- Numeric propositions (e.g. speed) should map to state-query functions.
- If a proposition requires internal state that is not directly observable from the provided APIs, introduce a global variable to maintain the state consistently.
- Low-level APIs should only be called inside proposition-aligned wrapper functions.
- Preserve exact naming alignment between propositions and generated APIs whenever possible.

Also generate:

1. A reusable helper layer if needed (e.g. set_velocity)
2. A Python plan template demonstrating safe execution of the skill
3. Runtime assertions enforcing global safety constraints

Output format:

## Skill: <skill_name>

### Global Contract

Logical formula

### Local Contract

Logical formula

### Semantic Contract

#### Proposition-aligned APIs

Example: python APIs

#### Plan Template

Example: python program
\end{lstlisting}

A user needs to provide a set of APIs or control commands and a set of global logical specifications. We present the control commands and global specifications for the Jackal ground robot and drone systems below.

\begin{lstlisting}[language=Python, style=code]
=====Jackal (Ground Robot)======
# APIs
def velocity_publisher(linear, angular):
    """Send linear and angular velocity commands."""
def yolo_listener(object_name) -> bool:
    """Return True if the specified object is observed."""

# Specifications
G(people_observed -> X stop)
G(obstacle_observed -> X not forward)
\end{lstlisting}

\begin{lstlisting}[language=Python, style=code]
======PX4 Drone (Aerial Robot)======
# Control Commands
A sequence of NED velocity in meters per second, published at 1 FPS.

# Specifications
G(max_altitude)
G(linear_velocity)
\end{lstlisting}

\paragraph{Convert Plan to Automaton}
Recall that we convert a generated plan into an automaton-based representation for model checking purpose, as described in Sec. \ref{sec:verification}. We present the algorithm of such conversion in Alg. \ref{alg:plan-to-automaton}. As we use NuSMV model checker, the automata are described in NuSMV coding format. We show several examples of the automata in NuSMV format and their corresponding python program in Tab. \ref{tab:nusmv_python}.

\newpage

\begin{algorithm}[t]
\caption{Plan-to-Automaton}
\label{alg:plan-to-automaton}
\begin{algorithmic}[1]
\REQUIRE Generated plan \(\pi\), proposition-aligned labeling function \(L_{sk}\)
\ENSURE Transition system \(\mathcal{A}_{\pi} = (S,S_0,T,L)\)

\STATE Initialize states \(S \leftarrow \emptyset\), transition \(T \leftarrow \emptyset\), and labeling function \(L\)
\STATE Extract the ordered executable sequence from \(\pi\):
\[
\sigma = \sigma_0,\sigma_1,\dots,\sigma_H
\]
where each \(\sigma_t\) is an API call, control command, waypoint, or trajectory segment

\FOR{\(t = 0,\dots,H\)}
    \STATE Create a state \(s_t\) corresponding to execution element \(\sigma_t\)
    \STATE Add \(s_t\) to \(S\)
    \STATE Assign symbolic labels using the proposition-aligned labeling function:
    \[
    L(s_t) \leftarrow L_{sk}(\sigma_t)
    \]
\ENDFOR

\STATE Set the initial state set \(S_0 \leftarrow \{s_0\}\)

\FOR{\(t = 0,\dots,H-1\)}
    \STATE Add transition \((s_t,s_{t+1})\) to \(T\)
\ENDFOR

\STATE \RETURN \(\mathcal{A}_{\pi} = (S,S_0,T,L)\)
\end{algorithmic}
\end{algorithm}

\begin{table}[t]
\centering
\small
\caption{Examples of plans composed from proposition-aligned APIs and their corresponding automata in \textsc{NuSMV}.}
\label{tab:nusmv_python}

\setlength{\tabcolsep}{6pt}
\renewcommand{\arraystretch}{1.15}

\begin{tabular}{>{\raggedright\arraybackslash}p{0.47\linewidth}
                >{\raggedright\arraybackslash}p{0.4\linewidth}}
\toprule
\textbf{\textsc{NuSMV}} & \textbf{Python} \\

\midrule
\begin{minipage}[t]{\linewidth}
\ttfamily
next(act) :=\\
\hspace*{1em}case\\
\hspace*{2em}TRUE : stop;\\
\hspace*{1em}esac;
\end{minipage}
&
\begin{minipage}[t]{\linewidth}
\ttfamily
while True:\\
\hspace*{1em}stop()
\end{minipage}
\\

\midrule
\begin{minipage}[t]{\linewidth}
\ttfamily
next(act) :=\\
\hspace*{1em}case\\
\hspace*{2em}speed < 1: stop;\\
\hspace*{2em}TRUE : forward;\\
\hspace*{1em}esac;
\end{minipage}
&
\begin{minipage}[t]{\linewidth}
\ttfamily
while True:\\
\hspace*{1em}if speed()<1:\\
\hspace*{2em}stop()\\
\hspace*{1em}else:\\
\hspace*{2em}forward()
\end{minipage}
\\

\bottomrule
\end{tabular}
\end{table}

\section{Experimental Details}
\label{app:experiment}

Fig. \ref{fig: testing} shows the safety scores of all specifications, where we present the full list of specifications below. In particular, the global specifications are evaluated across all the 400 plans, the local contracts are only evaluated via the plans generated from the particular skill (where we generate 40 plans for each skill).

\begin{lstlisting}[language=Python, style=code, caption={A set of specifications we verified during empirical evaluation.}, label={lst:spec}]
=====Jackal (Ground Robot)======
# Global Contract
1. G(people_observed -> X stop)
2. G(obstacle_observed -> X ! forward)
3. G(speed < 0.5)

# Local Contract
4. F !people_observed & !obstacle_observed -> F !stop
5. F obstacle_observed -> turn_left | turn_right

=====PX4 (Drone)======
# Global Contract
6. G(max_altitude < 10)
7. G(velocity_change <= 1)
8. G(linear_velocity <= 1)

# Local Contract
9. F (linear_velocity == 0 & altitude == 0)
10. G (altitude == 0 -> linear_velocity == 0)
11. F (linear_velocity > 0)
\end{lstlisting}

We show the prompts VLM-based planners under identical system specifications and temporal logic constraints. 

Listing~\ref{lst:baseline-prompts} directly instructs the model to generate executable plans while satisfying the safety specifications. 

\begin{lstlisting}[language={}, caption={Prompt exampkes for the baseline LLM-based planners.}, label={lst:baseline-prompts}]
Generate a sequence of NED velocity commands in m/s at 1 FPS to fly a 4 m x 4 m square trajectory.
- Commands are published once per second.
- Output only the velocity sequence, one command per line as: [vn, ve, vd]

Safety constraints:
- Always satisfy: max_altitude < 10 m
- Always satisfy: linear_velocity <= 1 m/s
- Always satisfy: velocity_change <= 1 m/s between consecutive commands

Task:
- Fly a square with side length 4 m.

Return exactly this kind of sequence:
[1, 0, 0]
[0, 1, 0]
[-1, 0, 0]
[0, 0, 0]
\end{lstlisting}

Listing~\ref{lst:react-prompt} extends the baseline by introducing an internal reasoning loop that explicitly performs observation and constraint verification before action generation. Note that Listing \ref{lst:react-prompt} only contains the additional components appended to the baseline prompt. 
\begin{lstlisting}[language={}, caption={Additional components for ReAct planner.}, label={lst:react-prompt}]
Internal ReAct loop:
1. Observe the current state, goal, obstacles, and previous command.
2. Reason about the next safest subgoal.
3. Check all hard constraints before choosing an action.
4. Act by outputting a valid velocity command.

Planning rules:
- Never violate constraints to complete the task.
- Prefer [0,0,0] when uncertain, blocked, or missing critical information.
- Keep vd = 0 unless altitude change is required.
- Do not climb if altitude is close to 10 m.
- Reduce speed near obstacles or targets.
- Insert [0,0,0] or intermediate velocities before large direction changes.
- Avoid oscillation by maintaining progress toward the current subgoal.
- Replan every step using the latest observation.
- If the task is complete, output [0,0,0].
\end{lstlisting}

Finally, for LLM+P, we augment the planner with a lightweight symbolic planning domain (Listing \ref{lst:llmp-domain}) whose predicates directly correspond to the temporal logic propositions in our specifications, enabling high-level task decomposition into safe executable actions.

\begin{lstlisting}[language={}, caption={Problem domain for LLM+P.}, label={lst:llmp-domain}]
(define (domain drone_velocity_safety)
  (:requirements :strips :typing)
  (:types
    drone velocity
  )
  (:predicates
    (max_altitude_lt_10 ?d - drone)
    (linear_velocity_le_1 ?v - velocity)
    (velocity_change_le_1 ?v_prev ?v_next - velocity)

    (current_velocity ?d - drone ?v - velocity)
    (commanded ?d - drone ?v - velocity)
  )
  (:action apply_velocity
    :parameters (?d - drone ?v_prev - velocity ?v_next - velocity)
    :precondition (and
      (current_velocity ?d ?v_prev)

      ;; Safety specs as predicates
      (max_altitude_lt_10 ?d)
      (linear_velocity_le_1 ?v_next)
      (velocity_change_le_1 ?v_prev ?v_next)
    )
    :effect (and
      (not (current_velocity ?d ?v_prev))
      (current_velocity ?d ?v_next)
      (commanded ?d ?v_next)
    )
  )
)
\end{lstlisting}

\subsection{Held-Out Skill Reuse Task Prompts}
\label{app:heldout-prompts}

To evaluate skill reuse, we optimize each skill on a set of in-domain prompts and evaluate the frozen optimized skill on held-out out-of-domain prompts without additional refinement. The out-of-domain prompts preserve the same underlying skill but introduce unseen parameterizations, compositions, and wording variations.

\begin{lstlisting}[language={}, caption={In-domain and held-out prompts for the \texttt{drive} skill (\texttt{Jackal}).}, label={lst:drive-prompts}]
Skill: drive

In-domain prompts:
1. Drive straight for 3 meters and stop.
2. Move forward 5 meters, then turn left by 90 degrees.
3. Travel 4 meters down the hallway and stop at the doorway.
4. Move forward for 2 meters and rotate right by 45 degrees.
5. Navigate 6 meters ahead while maintaining a straight path.

Out-of-domain prompts:
1. Follow an L-shaped path: move 3 meters forward, turn right 90 degrees, then move 2 meters.
2. Drive in a square pattern with side length 2 meters and stop at the starting position.
3. Go 5 meters forward, make a left turn, and continue another 3 meters.
4. Move to the hallway entrance, then rotate left to face the kitchen.
5. Traverse a zig-zag route by alternating 45-degree left and right turns every 2 meters.
\end{lstlisting}

\begin{lstlisting}[language={}, caption={In-domain and held-out prompts for the \texttt{find-object} skill (\texttt{Jackal}).}, label={lst:find-prompts}]
Skill: find-object

In-domain prompts:
1. Search for a chair and stop once it is detected.
2. Find a backpack near the hallway.
3. Navigate until a bottle is observed.
4. Search the room for a trash bin and stop when found.
5. Move through the environment until a person is detected.

Out-of-domain prompts:
1. Locate a chair positioned near the doorway and stop 1 meter away.
2. Search for a bottle while navigating around obstacles.
3. Move through the corridor until a backpack is found.
4. Find a person and stop immediately after detection.
5. Search for a trash bin while avoiding furnitures.
\end{lstlisting}

\begin{lstlisting}[language={}, caption={In-domain and held-out prompts for the \texttt{bypass-object} skill (\texttt{Jackal}).}, label={lst:bypass-prompts}]
Skill: bypass-object

In-domain prompts:
1. Move forward for 5 meters while avoiding obstacles.
2. Navigate to the doorway and bypass chairs blocking the path.
3. Travel 4 meters while safely avoiding detected obstacles.
4. Continue moving straight and reroute if an obstacle appears.
5. Reach the hallway entrance while avoiding chairs or boxes.

Out-of-domain prompts:
1. Move 6 meters forward while bypassing chairs.
2. Navigate around a chair and continue to move ahead.
3. Avoiding the chair on either the left and right sides.
4. Continue driving while avoiding furnitures ahead.
5. Rerouting around tables when necessary.
\end{lstlisting}

\begin{lstlisting}[language={}, caption={In-domain and held-out prompts for the \texttt{fly-in-square-shape-traj} skill (\texttt{PX4}).}, label={lst:fly-prompts}]
Skill: fly-in-square-shape-traj

In-domain prompts:
1. Fly in a square trajectory with side length 2 meters at altitude 3 meters and speed 0.5 m/s.
2. Execute a clockwise square flight with side length 3 meters while maintaining 5 meters altitude and velocity below 0.8 m/s.
3. Generate a square path of width 4 meters at altitude 4 meters using speed 0.6 m/s.
4. Fly a square route with side length 2.5 meters at 6 meters altitude and 0.7 m/s speed.
5. Complete a square-shaped trajectory at altitude 5 meters and speed 0.9 m/s.

Out-of-domain prompts:
1. Trace a square pattern with side length 5 meters at 7 meters altitude while keeping speed under 0.8 m/s.
2. Fly counterclockwise around a square with width 3 meters at altitude 4 meters using 0.5 m/s velocity.
3. Perform a square trajectory with a 1-second pause at each corner, altitude 6 meters, and speed 0.6 m/s.
4. Execute a square route with side length 2 meters at 8 meters altitude while limiting motion to 0.7 m/s.
5. Starting from hover, complete a square-shaped flight of width 4 meters at altitude 5 meters with maximum speed 0.9 m/s.
\end{lstlisting}

In particular, each skill is associated with 5 in-domain prompts used during optimization and 5 out-of-domain prompts reserved exclusively for testing. To account for the stochasticity of foundation model planning, we generate 5 plans per prompt using $\mathcal{F}_{\text{plan}}$ and evaluate each plan independently for specification satisfaction and task completion. Across the four skills, this results in a total of \(4 \times (5 + 5) \times 5 = 200\) evaluated plans, 50 per skill, where all out-of-domain results are obtained using the frozen optimized skill representation without additional refinement.

\newpage

\begin{table}[t]
\centering
\caption{Table of notations.}
\vspace{0.5em}
\begin{tabular}{lp{9cm}}
\toprule
\textbf{Notation} & \textbf{Description} \\
\midrule

\(\tau\) & Task prompt provided by the user. \\

\(\mathcal{AP}\) & Set of atomic propositions. \\

\(\sigma = \sigma_0, \sigma_1, \dots\) & Execution trace over time. \\

\(\sigma_t\) & Execution state at time \(t\). \\

\(sk\) & Skill. \\

\(\mathcal{G} = \{\varphi_1,\dots,\varphi_m\}\) & Set of global LTL specifications. \\

\(\psi_{sk}\) & Skill-specific local rule. \\

\(\mathcal{C}_{sk}\) & Semantic contract associated with a skill. \\

\(L_{sk}\) & Proposition-aligned labeling function. \\

\(\mathcal{C}_{sk}^{p}\) & Text-based plan template in the semantic contract. \\

\(\mathcal{F}_{\text{skill}}\) & Foundation model for skill generation and refinement. \\

\(\mathcal{F}_{\text{plan}}\) & Foundation model planner for executable plan generation. \\

\(\pi\) & Executable plan generated by the planner. \\

\(\mathcal{A}_{\pi}=(S,S_0,T,L)\) & Transition system compiled from a generated plan. \\

\(\mathcal{M}\) & Universal transition system representing all possible transitions. \\
\bottomrule
\end{tabular}
\label{tab:notations}
\vspace{-0.5em}
\end{table}

\section{Case Study}
\label{app: case}

\paragraph{Ground robot navigation.}
First, consider a simple navigation task on the Jackal platform: \textit{move forward for 5 meters and turn left by 90 degrees}. The framework initially generates a \texttt{drive} skill whose local rule is infeasible to the global specifications. Hence, we pass a feedback to the skill generation model $\mathcal{F}_{skill}$ to refine the local rule.
\begin{lstlisting}[caption={An example of feedback-guided skill regeneration.}, label={lst:regen}]
Original local rule:
G(!obstacle observed | person observed -> X forward)

Global specification:
G(person observed -> X stop)

Feedback: "The local rule conficts with the global specification G(person observed -> X stop)."

Regenerated local rule:
G(! (obstacle observed | person observed) -> X forward)
\end{lstlisting}
Then, the skill passes the skill-level verification, indicating the consistency between the global specifications and the skill-specific rule.

In addition to the global and local rules, the key component of the skill is the semantic contract, which consists of two components, the proposition-aligned labeling function and the plan template. In particular, the proposition-aligned labeling function is presented below.
\begin{lstlisting}[language=Python, style=code, caption={Proposition-aligned mappings for the Jackal ground robot.}, label={lst:prop-align}]
MAX_SPEED = 0.5
CURRENT_SPEED = 0.0
def people_observed() -> bool:
    return yolo_listener("people")
def obstacle_observed() -> bool:
    return yolo_listener("obstacle")
def speed() -> float:
    return CURRENT_SPEED

def set_velocity(linear: float, angular: float) -> None:
    global CURRENT_SPEED
    CURRENT_SPEED = abs(linear)
    velocity_publisher(linear=linear, angular=angular)
    
def forward() -> None:
    set_velocity(linear=0.3, angular=0.0)
def backward() -> None:
    set_velocity(linear=-0.3, angular=0.0)
def stop() -> None:
    set_velocity(linear=0.0, angular=0.0)
def turn_left() -> None:
    set_velocity(linear=0.0, angular=0.5)

def turn_right() -> None:
    set_velocity(linear=0.0, angular=-0.5)
\end{lstlisting}
Here, ``people\_observed()'' is aligned with the proposition ``people observed.''

The initial plan template in the semantic contract looks like
\begin{lstlisting}[language=Python, style=code]
while True:
    if people_observed() or obstacle_observed():
        break
    else:
        forward()
stop()
\end{lstlisting}
We then pass the skill into the planner $\mathcal{F}_{\text{plan}}$ to produce an executable plan. However, the generated plan fails to satisfy the local rule $F(\neg \texttt{people\_observed} \wedge \neg \texttt{obstacle\_observed}) \rightarrow F(\neg \texttt{stop})$, as the robot remains in a stop state indefinitely once a person or obstacle is detected. 

We then apply our \ours optimization to refine the semantic contract. After 10 iterations of refinement, the optimized skill satisfies all specifications and enables correct task execution. The optimized plan template is presented below.
\begin{lstlisting}[language=Python, style=code]
while True:
    if people_observed():
        stop()
    elif obstacle_observed():
        stop() # alternatively: turn_left() or turn_right()
    else:
        forward()
\end{lstlisting}

Additional examples of optimized skills and executions on the Jackal system are provided in Fig. \ref{fig:jackal-exe}.


\paragraph{Aerial navigation.}
Second, we consider a task on a PX4 quadcopter: \textit{fly in a square-shaped trajectory}. Since the system commands for the quadcopter are low-level velocity trajectories, i.e., list of NED tuples, the proposition alignment is more complicated than Jackal's proposition-API mappings. We present partial mapping below.
\begin{lstlisting}[language=Python, style=code, caption={Proposition-aligned mappings for the drone.}, label={lst:prop-align-drone}]
Velocity = Tuple[float, float, float]

def _norm3(v: Velocity) -> float:
    return math.sqrt(v[0] ** 2 + v[1] ** 2 + v[2] ** 2)

def _clip_norm(v: Velocity, max_norm: float) -> Velocity:
    n = _norm3(v)
    if n == 0.0 or n <= max_norm:
        return v
    s = max_norm / n
    return (v[0] * s, v[1] * s, v[2] * s)

def _altitude_trace(plan: List[Velocity], initial_altitude: float = 0.0, 
        dt: float = 1.0) -> List[float]:
    trace = [initial_altitude]

    for _, _, vd in plan:
        z += vd * dt
        trace.append(z)

    return trace

def linear_velocity(v: Velocity) -> float:
    return _norm3(v)

def velocity_change(previous: Velocity, current: Velocity) -> float:
    return _norm3((
        current[0] - previous[0],
        current[1] - previous[1],
        current[2] - previous[2],
    ))

def altitude(plan: List[Velocity], idx=-1) -> float:
    return _altitude_trace(plan)[idx]
\end{lstlisting}
We want to highlight that our proposition alignment can include numerical computations, which \textbf{maps low-level continuous signals to high-level propositions.}

Then, the plan template guides the planner to generate a list of NED velocities, which can be directly sent to the quadcopter system.
\begin{lstlisting}[language=Python, style=code]
vel_traj: List[Velocity] = []
ver_speed, vert_duration_s = 1, 10

vel_traj += [(0.0, 0,0, -ver_speed)] * vert_duration_s  # take off
vel_traj += [(0.0,  speed_mps, 0.0)] * side_duration_s  # east
vel_traj += [(-speed_mps, 0.0, 0.0)] * side_duration_s  # south
vel_traj += [(0.0, -speed_mps, 0.0)] * side_duration_s  # west
vel_traj += [( speed_mps, 0.0, 0.0)] * side_duration_s  # north
vel_traj += [(0.0,  0,0, ver_speed)] * vert_duration_s  # land
\end{lstlisting}
The generated plan guided by the initial skill violates the global specification $G(\texttt{max\_altitude} < 10)$ due to a subtle edge case that can be easily overlooked by human inspection. In particular, when the altitude reaches 10 meters, the drone continues to ascend slightly until the next control command is issued, resulting in a small but critical violation of the safety constraint.

After one iteration of skill optimization, by only changing one line of code (as presented below), the refined skill satisfies all specifications and executes the task safely. 
\begin{lstlisting}[language=Python, style=code]
ver_speed, vert_duration_s = 0.5, 10
\end{lstlisting}

These two demonstrations illustrate how the framework systematically identifies and corrects both obvious and subtle violations through verification-guided skill optimization. In the ground navigation task, the framework resolves persistent liveness failures, while in the aerial setting it captures edge-case safety violations that are easily overlooked by human inspection. In both cases, a small number of refinement iterations suffices to obtain skills that satisfy all specifications and enable reliable execution.

\paragraph{Failure Case.}
Our framework replies on the proposition-aligned labeling function correctly maps low-level robot actions to atomic propositions. 
When this assumption is violated, formal verification may produce false safety guarantees. 
For example, consider the PX4 specification
\[
\mathbf{G}(\texttt{linear\_velocity} < 1),
\]
which constrains the drone's velocity magnitude. 
The proposition-aligned mapping is incorrectly implemented as:
\begin{lstlisting}[language=Python, style=code]
def _norm3(v: Velocity) -> float:
    return math.sqrt(v[0] ** 2 + v[1] ** 2)

def linear_velocity(plan: List[Velocity], idx=-1) -> float:
    return _norm3(plan[idx])
\end{lstlisting}
Here, the implementation mistakenly computes only the horizontal velocity in the north-east plane and ignores the vertical velocity component in the NED frame. 
As a result, a velocity command such as
\[
(0.8,\;0.6,\;-0.5)
\]
is incorrectly evaluated as
\[
\sqrt{0.8^2 + 0.6^2} = 1.0,
\]
which satisfies the specification threshold in automated verification, while the true linear velocity magnitude is
\[
\sqrt{0.8^2 + 0.6^2 + (-0.5)^2}
= 1.12 > 1,
\]
thus violating the specification. 

When verified using a handcrafted proposition-aligned labeling function that correctly accounts for all NED velocity components, the same plan fails verification. 
This example highlights that the correctness of verification fundamentally depends on the fidelity of the proposition-aligned mapping; an incorrect abstraction of low-level robot dynamics may lead to spurious verification results despite formally correct reasoning over the symbolic model.

Despite this limitation, we empirically observe that such proposition misalignment errors occur infrequently in practice. 
Across the \ours optimization process, comparing auto-generated proposition-aligned mappings against handcrafted implementations results in less than a \(1\%\) degradation in the final safety score. 
This suggests that, although imperfect proposition grounding may occasionally introduce spurious verification outcomes, the overall verification procedure still provides sufficiently accurate feedback to effectively guide skill optimization. 

A future direction is to integrate formal verification with \emph{probabilistic grounding validation} to improve robustness against low-level abstraction errors. 
Instead of treating proposition-aligned mappings as deterministic functions, future systems may associate each proposition with a confidence estimate that quantifies the reliability of its grounding from low-level robot states.

\section{Ablation Study}
\label{app:ablation}
We isolate the contribution of three core design choices in \ours: structured \texttt{skill.MD}, automatic proposition alignment, iterative skill optimization, and skill-mediated planning.

\textbf{Structured Skill Representation.}
Fig. \ref{fig: testing} compares \ours against LLM-AutoDiff \cite{yin2025llm-autodiff}, which optimizes flat natural-language planning instructions without any structured skill representation. \ours surpasses 90\% safety score within 7 training steps; LLM-AutoDiff plateaus near 85\% after 10 steps. The difference in both convergence speed and final performance shows that the structured, verifiable representation is the key driver of improvement, not additional optimization steps.

\textbf{Automatic Proposition Alignment.}
We query $\mathcal{F}_{skill}$ to generate proposition-aligned labeling function that maps low-level control commands to propositions. However, this auto-generated function may not always provide correct mapping. Hence, we provide handcrafted proposition alignment functions as ground truth and for comparison. Tab. \ref{tab:planning-baselines} shows that the auto-labeling achieves on-par performance with manual-labeling, while eliminating the need for human experts.

\textbf{Iterative Skill Optimization.}
Fig. \ref{fig: testing} (right) shows that plans generated from initial, unrefined skills achieve substantially lower and more variable safety scores compared to plans from optimized skills, which consistently reach 95\% compliance across all 11 specifications and both platforms. The large and consistent gap across diverse constraints shows the necessity of the iterative refinement.

\textbf{Skill-Mediated Planning.}
Tab. \ref{tab:planning-baselines} shows that removing skill mediation in favor of direct end-to-end, whether zero-shot or learning-based, consistently, reduces safety compliance. The strongest learning-based baseline, RLVF~\cite{yang-mlsys}, reaches 93.5\% safety score while \ours achieves 97.2\%. This highlights that structuring the planning interface through verified skills yields better safety compliance with substantially less computational cost.